\documentclass[runningheads]{llncs}

\usepackage{eccv}

\usepackage{eccvabbrv}
\usepackage{diagbox}
\usepackage{graphicx}
\usepackage{booktabs}
\usepackage{multirow}
\usepackage[symbol]{footmisc}
\usepackage[accsupp]{axessibility}  
\definecolor{green}{HTML}{00BC12}

\usepackage[pagebackref,breaklinks,colorlinks,citecolor=eccvblue]{hyperref}

\usepackage{orcidlink}

\begin{document}

\title{Rethinking Clothes Changing Person ReID: Conflicts, Synthesis, and Optimization} 

\titlerunning{Rethinking Clothes Changing Person ReID} 
\authorrunning{J. Li et al.} 
\author{Junjie Li$^{1}$\thanks{Equal contribution, This work was done during Li’s internship at Tencent Youtu.}, Guanshuo Wang$^{2*}$, Fufu Yu$^2$, Yichao Yan$^{1}$\thanks{Corresponding author: Yichao Yan.}, Qiong Jia$^2$,\\ Shouhong Ding$^2$, Xingdong Sheng$^3$, Yunhui Liu$^3$, Xiaokang Yang$^1$}
\institute{$^1$MoE Key Lab of Artificial Intelligence, AI Institute, Shanghai Jiao Tong University \quad $^2$Tencent Youtu Lab \quad $^3$ Lenovo \\
{\tt\small \{junjieli00,yanyichao,xkyang\}@sjtu.edu.cn, \\ 
\{mediswang,fufuyu,boajia,ericshding\}@tencent.com, \\
\{shengxd1,liuyhp\}@lenovo.com
}}

\maketitle

\begin{abstract}
Clothes-changing person re-identification (CC-ReID) aims to retrieve images of the same person wearing different outfits. Mainstream researches focus on designing advanced model structures and strategies to capture identity information independent of clothing. However, the same-clothes  discrimination as the standard ReID learning objective in CC-ReID is persistently ignored in previous researches.
In this study, we dive into the relationship between standard and clothes-changing~(CC) learning objectives, and bring the inner conflicts between these two objectives to the fore.
We try to magnify the proportion of CC training pairs by supplementing high-fidelity clothes-varying synthesis, produced by our proposed Clothes-Changing Diffusion model. By incorporating the synthetic images into CC-ReID model training, we observe a significant improvement under CC protocol.
However, such improvement sacrifices the performance under the standard protocol, caused by the inner conflict between standard and CC.
For conflict mitigation, we decouple these objectives and re-formulate CC-ReID learning as a multi-objective optimization (MOO) problem. By effectively regularizing the gradient curvature across multiple objectives and introducing preference restrictions, our MOO solution surpasses the single-task training paradigm. Our framework is model-agnostic, and demonstrates superior performance under both CC and standard ReID protocols.
\end{abstract}

\textbf{\textsl{Conflict is inevitable, but combat is optional.}}\hfill \textbf{\textsl{- Max Lucado}}

\section{Introduction}
\label{sec:intro}
\begin{table}[t]
\caption{Mean Average Precision (mAP) comparison by different training paradigm and test protocols on the PRCC dataset. `CC' means the pure changing clothes protocol, and `SC' means the pure same clothes protocol.}
\setlength{\abovecaptionskip}{2mm}
\centering
\begin{tabular}{l|p{2.5cm}<{\centering}|p{2.5cm}<{\centering}}
\hline
\diagbox{Training}{Test}  &\multicolumn{1}{c|}{CC} &\multicolumn{1}{c}{SC}
\\ \hline
Original  &45.4\% &99.1\% \\ \hline
\hline
SC (label transformation)  &$27.8\%_{(\textcolor{red}{-17.6\%})}$ &$99.9\%_{(\textcolor{green}{+0.8\%})}$ \\ \hline
\hline
CC (w/ synthesis) &$61.7\%_{(\textcolor{green}{+16.3\%})}$ &$89.4\%_{(\textcolor{red}{-9.7\%})}$ \\ 
\hline
CC (w/ synthesis+GBO) &$63.8\%_{(\textcolor{green}{+18.4\%})}$ &$97.1\%_{(\textcolor{red}{-2.0\%})}$ \\ 
\hline
\end{tabular}
\label{toy-sc}
\end{table}

Person re-identification~(ReID) is a task that aims to retrieve images depicting the same person from a given gallery set. Standard/Same-clothes~(SC) ReID methods have predominantly relied on the assumption that individuals maintain clothing consistency across different views, treating clothing information as an integral part of identifiable features. However, this assumption does not hold in the context of the recently proposed clothes-changing person ReID (CC-ReID) setting, which has garnered increasing attention due to its crucial role in public security systems.
standard ReID approaches often struggle to handle scenarios involving changes in clothing, prompting extensive research efforts in this direction~\cite{DBLP:conf/cvpr/HongWWHZ21, DBLP:conf/cvpr/GuCMBS022, DBLP:conf/cvpr/YangLZW023, han2023clothing}.
State-of-the-art approaches in the field of CC-ReID have been dedicated to enhancing clothes-changing~(CC) performance through designing specialized model structures and learning strategies aimed at effectively separating clothing information from identity representation.

A single person in typical CC-ReID datasets would wear different styles of clothes. For each clothing, the dataset contains several sample images showcasing the individual wearing it. Therefore, the learning process involve both standard ReID~(same-clothes instances as positive samples) and CC~(instances with different clothes as positive samples) objectives.
However, the standard ReID training objective in CC-ReID is persistently \textbf{ignored} in previous CC-ReID studies.
standard ReID assumes that one identity does not change clothes, thus clothing is the most dominant feature, and serves as the primary ID-related characteristic. Conversely, in the CC context, the assumption about unchanging clothes does not hold. This leads to the focus being shifted to other clues like face, hair, and body shape. Intuitively, such discrepancy about reliance on different visual cues would lead to \textbf{conflicts} between learning objectives of standard and clothes-changing ReID.
To investigate this, we conducted a toy experiment to increase the percentage of SC/standard ReID training pairs in CC-ReID. More specifically, this is achieved by treating instances of the same person wearing different clothes as different identities~(transforming identity annotation according to clothes). The result is reported in `SC~(label transformation)' row in Tab~\ref{toy-sc}. While the accuracy on the SC test set improves from 99.1\% to 99.9\%, a clear decrease~(\textcolor{red}{17.6\%}) in CC-ReID performance can be observed. The results of this toy experiment not only presents that increasing the percentage of standard ReID learning cases is beneficial for SC performance, but also validates the conflicts between standard and CC ReID learning objectives.
The increase of same-clothes pairs in this toy experiment eventually enhances the standard ReID learning. Considering the contradicting relation between standard and CC ReID objectives, one question is naturally raised: \textbf{\textsl{Can we improve the Clothes-changing performance by increasing the proportion of CC training pairs in CC-ReID?}}

A naive solution to achieving this is extending clothes diversity of existing dataset, which means collecting more real-world clothes changing data with identities of existing dataset. However, the difficulty and cost of ensuring identity consistency on existing dataset hampers the progress in this direction. Some efforts~\cite{zheng2019joint,DBLP:journals/spl/ShuLWRT21,han2023clothing} have been dedicated to exploring clothing-related augmentation, they either focus on enhancing standard ReID task~\cite{zheng2019joint}, or concentrate on basic pixel/feature transformations~\cite{DBLP:journals/spl/ShuLWRT21,han2023clothing}.
Due to inconsistency and low controllability, these methods are not comparable to real-world collected data.
Recently, diffusion models \cite{ho2020denoising, song2020denoising, liu2022pseudo, rombach2022high, zhu2023tryon, kim2023dcface} gradually become a mainstream image synthesis method due to its excellent generation quality and flexible controllability. In this paper, we introduce high-fidelity synthesized clothes-varying person images via our proposed Clothes-Changing Diffusion model. CC-Diffusion takes different clothes in the same datasets as controlling conditions, and synthesizes clothes-varying samples with consistent physical features from given persons. To better handle the clothes warping between given conditions and person wearing, CC-Diffusion consists of two branches of UNet to respectively handle clothes warping and person image rendering. Quantitative experiments show the high synthesis quality and effective improvement on CC-ReID tasks.

However, considering the discrepancy between SC and CC ReID objective, introducing CC synthesis inevitably shifts the focus towards cloth-irrelevant clues, and weakens the standard ReID objective. The results presented in row `CC (w/ our synthesis)' of Tab~\ref{toy-sc} highly corresponds to such trend, that our CC synthetic data significantly boosts the by \textcolor{green}{16.3\%}, but the SC performance drops severely by \textcolor{red}{9.7\%}. This result further validates the contradicting learning objectives in CC-ReID, and additional clothes-changing data amplifies such conflicts. Therefore, another question is naturally raised: \textbf{\textsl{Can we mitigate the conflicts between standard and CC objectives in CC-ReID training?}}

Considering the inner conflicts in CC-ReID training, the common practice~\cite{DBLP:conf/cvpr/GuCMBS022,han2023clothing,DBLP:journals/spl/ShuLWRT21, DBLP:conf/mm/GuoLWWW23} of training within a integrated CC-ReID dataset can be viewed as a naive strategy of linearly weighting these objectives according to inherent data proportions. However, due to the non-convex~\cite{boyd2004convex} property, such a practice does not necessarily yield optimal results for both sub-objectives.
To mitigate the conflicts and achieve optimal trade-offs, the conflicting objectives ought to be disentangled and optimized in a synergistic manner. By properly partitioning the training samples and designing the sampling strategies, we disentangle CC-ReID training into multiple competing optimization goals, representing standard and CC objectives. Following this line, we delve into techniques of multi-objective optimization~(MOO) and extend them to our decomposed training process to achieve a set of Pareto optimal solutions~(not be dominated by any other solution in the objective space). As is presented in Tab~\ref{toy-sc}, our proposed formulation not only further improves CC discrimination ability, but also narrows the SC performance drop by \textcolor{green}{7.7\%}.
To freely strike a desired balance between standard and CC-ReID rather than obtain a random pareto optimal solution, we further introduce preference vectors that capture human inclination. The effectiveness of our framework is validated on both standard and specialized CC-ReID models.

Our contributions are summarized as follows
\begin{itemize}
    \setlength{\itemsep}{2pt}
	\setlength{\parsep}{-3pt}
	\setlength{\parskip}{-0pt}
	\setlength{\leftmargin}{-15pt}
    \item To the best of our knowledge, we first uncover the inner conflicts between standard and clothes-changing learning objectives in CC-ReID.
    \item Inspired by the trade-offs in CC-ReID, we propose to synthesize high-fidelity clothes-changing data for enhancing CC performance for the first time. Our synthetic data demonstrates superior quality, diversity, and controllability.
    \item To mitigate the inner conflicts and achieve optimal solutions, we re-formulate the learning of CC-ReID as a multiple-objective optimization~(MOO) problem. Moreover, we introduce human preference vectors to ensure convergence to the desired balance. 
    \item Our framework is model-agnostic. When applied to a standard ReID model, our method surprisingly outperforms existing CC-ReID methods.
\end{itemize}

\section{Related Works}
\textbf{Clothes-Changing ReID.}
ReID tasks focus on capturing intricate details and distinctive characteristics of individual identities, and traditional methods tend to treat clothing as a static and consistent attribute. Based on this, an active research line~\cite{DBLP:conf/eccv/SunZYTW18, DBLP:conf/mm/WangYCLZ18, DBLP:conf/cvpr/HouMCGSC19a} is introducing fine-grained part features into representation to improve the robustness. Several works~\cite{DBLP:journals/corr/HermansBL17, DBLP:conf/cvpr/SunCZZZWW20,DBLP:conf/cvpr/XiaoLWLW17} also focus on developing metrics to enhance the representation learning. However, models trained with standard ReID data typically struggle to handle clothes changing cases due to the inductive bias about clothes.
With the significance of long-term identification receiving attention, several CC-ReID datasets~\cite{DBLP:journals/pami/YangWZ21, DBLP:conf/cvpr/GuCMBS022,DBLP:conf/accv/QianWZZFXJX20, DBLP:conf/cvpr/WanWQCF20,DBLP:conf/mm/WangYYSLL23} have been collected to facilitate clothes-invariant ReID systems development. Existing methods for CC-ReID can be categorized into two branches. The first branch~\cite{DBLP:conf/cvpr/YangLZW023,DBLP:conf/cvpr/GuCMBS022} utilizes the clothing labels to decorrelate clothing information from identity feature.
On the contrary, the other branch of works incorporate auxiliary modality information,~\emph{e.g.}, human parsing~\cite{DBLP:journals/spl/ShuLWRT21, DBLP:conf/mm/GuoLWWW23}, pose estimation~\cite{DBLP:conf/cvpr/HongWWHZ21, DBLP:conf/accv/WangQFX22}, and silhouettes~\cite{DBLP:conf/cvpr/HongWWHZ21,DBLP:conf/bmvc/MuLLY22,DBLP:conf/cvpr/JinHZYSHFH0022} into feature learning.
While most existing CC-ReID methods focus on designing model structures and developing clothes-independent feature learning, our work takes a different perspective by considering the data itself, and re-formulate the learning as simultaneously optimizing conflicting objectives.

\textbf{Diffusion-based Image Synthesis} In recent image synthesis tasks, diffusion models \cite{song2019generative,ho2020denoising,song2020denoising,nichol2021improved,song2021maximum,liu2022pseudo} achieves surprising synthesi quality, beating traditional GAN models \cite{kingma2021variational,dhariwal2021diffusion,rombach2022high}. Especially, DMs achieves success in hint-guided synthesis tasks such as text-to-image \cite{rombach2022high, saharia2022photorealistic, ruiz2023dreambooth, kawar2023imagic, kumari2023ablating, yang2023zero} and image-to-image \cite{kim2022diffface, yang2023paint, zhu2023tryon, kim2023dcface, boutros2023idiff} synthesis. With excellent synthesis quality, DMs are utilized for data augmentation \cite{trabucco2023effective, burg2023data} to improve generalization performance. Also inspired by the success of virtual tryon methods \cite{zhu2023tryon, gou2023taming}, it is reasonable to synthesize clothes-varying samples of certain person images by different clothes hint guidance, by which we can achieve significant improvement on CC-ReID tasks. 

\textbf{Multi-Objective Optimization.}
Multi-objective optimization aims to effectively address conflicting learning objectives by seeking optimal solutions. An intuitive practice is using pre-defined linear weights to balance multiple objectives. Alternatively, some works leverage gradient magnitude~\cite{chen2018gradnorm}, and task uncertainty~\cite{kendall2018multi} to automatically determine the balancing weights. Another line of works~\cite{desideri2012multiple, DBLP:journals/siamjo/FliegeV16,DBLP:conf/nips/SenerK18,mahapatra2020multi} focus on using gradient to search for a descent direction that dominates the previous status for all objectives. MGDA~\cite{desideri2012multiple} extended the gradient-based methods to the widely used SGD, and got validated by multi-task learning~\cite{DBLP:conf/nips/SenerK18,mahapatra2020multi, liang2021pareto}, multi-agent learning~\cite{ghosh2013towards, poirion2017descent}. In this work, we propose to formulate CC-ReID learning as a MOO problem, and search the desired pareto optimal solutions using the gradient-based method with human preference.

\section{Method}
\begin{figure}[t]
  \centering
  \includegraphics[width=\linewidth]{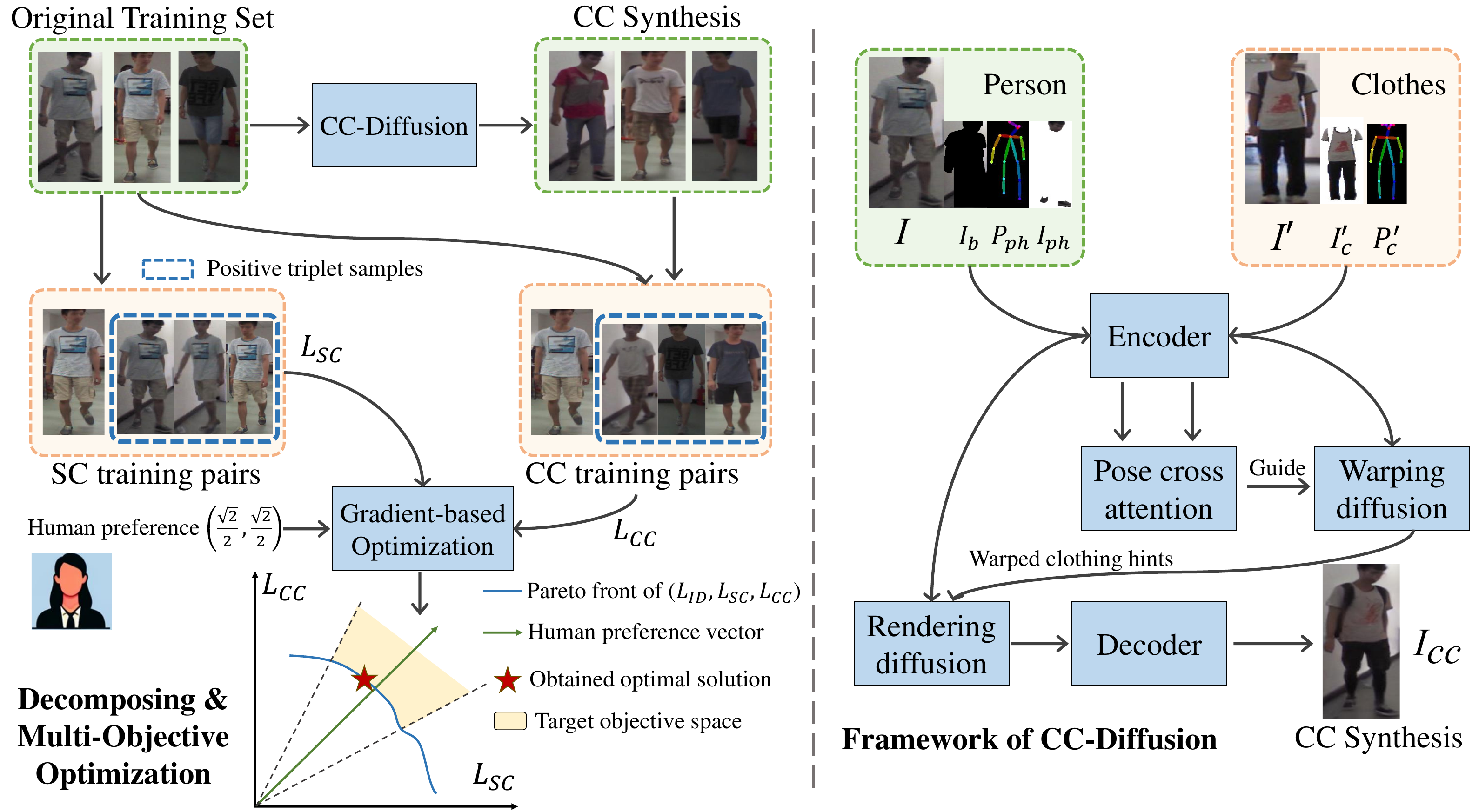}
  \caption{The overview of our framework and the pipeline of generating synthesis with Clothes-Changing Diffusion model. The learning process of standard and CC pairs are decomposed to jointly optimize for pareto-optimal solutions.} 
  \label{fig:pipeline}
\end{figure}
We present the overall framework of our method in Fig~\ref{fig:pipeline}. To investigate the first problem proposed in introduction, we explore Diffusion-based synthesis to increase the proportion of CC training pairs. To answer the second question about mitigating learning conflicts, we present a new formulation of decomposing objectives and optimizing multiple controversial objectives.

\subsection{Synthesizing Varying-Clothes Samples}
Fig~\ref{fig:pipeline} demonstrates the pipeline of our proposed Clothes-Changing Diffusion~(CC-Diffusion) model. In synthesis process, we intend to reserve essential environmental and identity-related feature of a reference image $I$ to ensure fidelity. So we first decompose this image into clothes parts $I_c$, physical parts $I_{ph}$, and background $I_{b}$ via human parsing~\cite{li2020self}, where $I_{ph}$ and $I_{b}$ are retained, but $I_c$ from are replaced with clothes images $I_c^{'}$ from another person. However, the segmented borders in condition images may cause severe constraints on the cross-style clothes changing. So we conduct dilation preprocessing to the condition images. Besides, to associate their spatial relationship, the estimated pose skeleton \cite{8765346} are used as side information of spatial positioning. We denote the original pose image as $P_{ph}$, and the pose of changed clothes as $P_c^{'}$.
\begin{figure}[t]
\setlength{\abovecaptionskip}{8pt}
  \centering
  \includegraphics[width=0.85\linewidth]{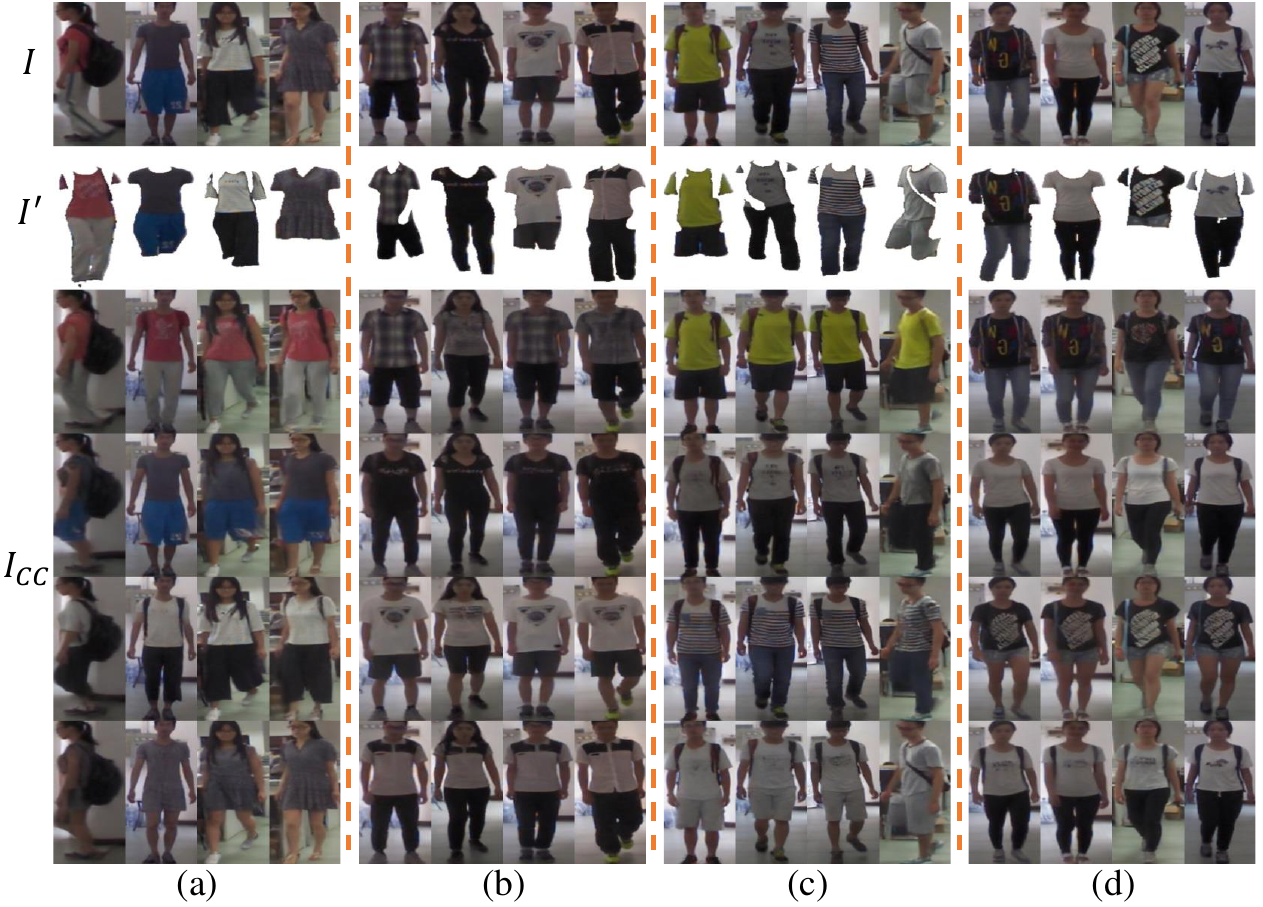}
  \caption{Synthetic results on PRCC, every 4 persons is grouped together. The first row contains original images $I$, the second row refers to clothes images $I'$. The other rows represent clothes-varying synthesis $I_{CC}$ by swapping clothing in the group.}
  \label{fig:visualization}
\end{figure}

As is illustrated in Fig~\ref{fig:pipeline}. The input sample is encoded into a latent representation with encoder. Latent representations can reconstruct the input images by the decoder. Condition images $I_c^{'}$, $I_b$ and $I_{ph}$ are all encoded into latent representations $z_c^{'}$, $z_b$ and $z_{ph}$, and side information $P_{ph}$ and $P_c^{'}$ are linearly projected into $p_{ph}$ and $p_c^{'}$ with the same spatial and channel sizes. The diffusion process consists of two processes. The forward process gradually corrupts the sample $x_t$ at timestep $t$ by noise $\epsilon$. The reverse process predicts the noise at timestep $t$ to denoise the sample according to the sample $x_t$ and clothes-changing conditions $C$ as $\epsilon_\theta(x_t, t, C)$. We conduct the diffusion process by a two-step pipeline. Clothing conditions $z_c^{'}$ first are warped into a fit shape condition for clothes wearing via a warping diffusion model. To give accurate hints of clothes warping, we design the cross attention between query target poses $p_{ph}$ and key/value source clothes poses $p_c^{'}$ as pose hint $c_{pose}=\text{Attention}(p_{ph}, p_c^{'}, p_c^{'})$. Then we apply this as the key/value context embedded to warping diffusion by cross attention. All the noised samples to reserved conditions and warped hints by the warping step are concatenated with the noise $x_t$ along the channel-axis, as the input of the rendering diffusion. The warping and rendering steps are employed for $T$ times to get the final denoised $\tilde{x}$, as the latent code for final synthetic image.
During training, the encoded latent representations of original images $z$ and conditions $C=\{z_{ph}, z_{b}, z_{c}^{'}\}$ are all corrupted by Gaussian noise $\epsilon\sim\mathcal{N}(0, I)$ for $T$ times. The input noise sample of the original images are denoted as $x_T$. The learning object is to restore the original images by denoising, and the conditional diffusion loss function is formulated as 
\begin{equation}
    L = E_{\mathcal{E}(x), \epsilon\sim\mathcal{N}(0, I), t}\left[ \| \epsilon-\epsilon_\theta(x_t, t, C) \|^2_2 \right],
\end{equation}
To match the clothes in original images, the changing condition $z_c^{'}$ in $C$ should be the latent representation from the clothes images in the same style as the original, but with different poses or orientation. 

In synthesis process, for each clothing in a identity, we randomly sample 5 different clothes within the dataset as changing conditions. As a result, we can obtain $5\times$ the amount of synthesis with different appearance but remained identity-related features.
We provide some synthetic results in Fig~\ref{fig:visualization}. Specifically, four randomly selected identities are grouped into a batch and swap their outfits. The synthetic results fully demonstrate the superiority of our CC-diffusion in preserving texture details in clothes~(group b, c, d), crossing clothing type and human pose~(group a, b), and keeping identity information~(group a, b, c, d).

\subsection{Optimizing Controversial Objectives}
Motivated by the toy experiment in the `SC (label transformation)' row of Tab~\ref{toy-sc}, that the increase of standard ReID pairs leads to an enhancement in SC performance, but at the expense of CC performance. This intriguing observation prompts us to hypothesize that the learning objectives of standard ReID and CC-ReID may conflict with each other. The conjecture is in line with intuition, as standard ReID places importance on clothing as a distinguishing factor, whereas CC-ReID considers it as a ID-irrelevant information. 

By applying our synthesis to training, the probability of obtaining clothes-changing training samples is expected to increase. When using the widely employed ${PK}$ sampling strategy, where ${P}$ identities and ${K}$ instances of each identity are sampled in sequence. Suppose one identity has $N_c$ clothes, the possibility of sampling clothes-changing examples is bound to increase when we apply enough synthetic data to make $N_c \leq k_c+1$~(please check Appendix for detailed proof).
The `CC (w/ our synthesis)' row in Tab~\ref{toy-sc} reflects the effect of adding the synthetic data to ReID training, that the performance on CC test sets is significantly boosted. However, a noticeable decline is observed on the same-clothes protocol. These results further validates the conflicts between CC and standard objectives. Based on these observations, we propose to decompose the learning of CC-ReID:
\begin{equation}
\begin{aligned}
&\mathcal{L}_{id} = \mathcal{L}_{i d}(x_{sc} ; \theta) + \mathcal{L}_{i d}(x_{cc} ; \theta) ,\\
\mathcal{L}_{sc} &= \mathcal{L}_{tri}(x_{sc} ; \theta) ,\quad
\mathcal{L}_{cc} = \mathcal{L}_{tri}(x_{cc} ; \theta) .
\end{aligned}
\end{equation}
where $x_{sc}$ and $x_{cc}$ represents standard and clothes changing training data, respectively. In practice, we transform the ${PK}$ sampler to construct clothes changing pairs for $\mathcal{L}_{cc}$. In the new sampler, $K$ instances from a given identity would wear $K$ different clothes. Similarly, $x_{sc}$ is obtained from a modified ${PK}$ sampler for constructing same clothing pairs. By using these samplers, we employ the triplet loss~\cite{DBLP:journals/corr/HermansBL17} to explicitly learn representations with and without the clothing factor. 

\textbf{Optimizing Decomposed CC-ReID Problem.}
With decomposition above, CC-ReID is transformed into a multi-objective optimization~(MOO) problem.
\begin{equation}
\begin{aligned}
\label{eq:moo}
\min _{\theta} \hat{\mathcal{L}}(\theta) = ( \mathcal{L}_{id} (\theta), \mathcal{L}_{sc} (\theta), \mathcal{L}_{cc} (\theta) )^\mathrm{T}
\end{aligned}
\end{equation}
An intuitively appealing solution to this problem is optimizing a linearly weighted composition of the learning goals. This solution involves manually setting weighting hyper-parameters in advance. Unfortunately, such weighting methods can hinder the optimization of certain objectives during training. Moreover, these methods can only yield solutions within the convex region of the Pareto front, failing to explore the non-convex part. In the case of CC-ReID, the conflicts among learning objectives make the Pareto front highly non-convex~\cite{boyd2004convex}. Consequently, relying on weighting methods would inevitably result in sub-optimal solutions that are dominated by other solutions.
To overcome these limitations, we follow the gradient-based methods by leveraging the gradients information to seek pareto optimal solutions. The overall optimization objective can be formulated as finding a solution $\boldsymbol{\theta^{*}}$ that would not be dominated by any other solution. In other words, there should be no solution $\boldsymbol{\theta}$ satisfying:
\begin{equation}
\begin{aligned}
&\hat{\mathcal{L}}^i(\boldsymbol{\theta}) \leq \hat{\mathcal{L}}^i(\boldsymbol{\theta^{*}}), \quad \forall i \in \{1, 2, \ldots, m\},\\
&\hat{\mathcal{L}}^i(\boldsymbol{\theta}) < \hat{\mathcal{L}}^i(\boldsymbol{\theta^{*}}), \quad \exists i \in \{1, 2, \ldots, m\},
\end{aligned}
\end{equation}
where $m=3$ denotes the number of objectives. The multi-objective optimization problem can be optimized to local optimal points via multiple gradient descent~\cite{desideri2012multiple, DBLP:journals/siamjo/FliegeV16}. In each training step, the solution aims to find a descent direction that improves all decomposed objectives through the following formulation:
\begin{equation}
\begin{aligned}
\label{eq:opt_goal}
& \min _{\alpha^1, \ldots, \alpha^m} \left\|\sum_{i=1}^m \alpha^i \nabla_{\theta} \hat{\mathcal{L}}^i(\theta)\right\|_2^2,
\text { s.t. } \sum_{i=1}^m &\alpha^i=1, \alpha^i \geq 0, \forall i \in \{1, 2, \ldots, m\},
\end{aligned}
\end{equation}
the descent direction in each iteration would lead to a solution dominating the previous one. When the formulation in~\ref{eq:opt_goal} is optimized to 0, the achieved optimal point would certainly satisfy a Karush-Kuhn-Tucker (KKT) condition:
\begin{equation}
\label{eq:p_stationary}
\exists \sum_{i=1}^m \alpha^i=1, \alpha^i \geq 0, \quad\sum_{i=1}^m \alpha^i \nabla_{\theta} \hat{\mathcal{L}}^i(\theta) = 0.
\end{equation}
Therefore, the gradient-based optimization approach can be viewed as a two-step adaptive weighting method. In the first step, we utilize the formulation~\ref{eq:opt_goal} to obtain adaptive weights $\alpha=\{\alpha^1, \ldots, \alpha^m\}$ for the current iteration. In the second step, the losses are balanced and weighted using $\alpha$.
The resulting parameters correspond to a pareto stationary solution, where the model cannot improve the performance of one decomposed goal without increasing the loss of another goal. Achieving pareto stationary point is a necessary condition for obtaining pareto optimal solution, but not a sufficient one. However, with the non-singularity assumption proposed in previous works~\cite{DBLP:conf/nips/SenerK18}, such methods can lead to a pareto optimal point. Fortunately, the non-singularity assumption is easily satisfied in our problem since the decomposed objectives are not linearly related.

\begin{figure}[t]
\setlength{\abovecaptionskip}{8pt}
  \centering
  \includegraphics[width=0.8\linewidth]{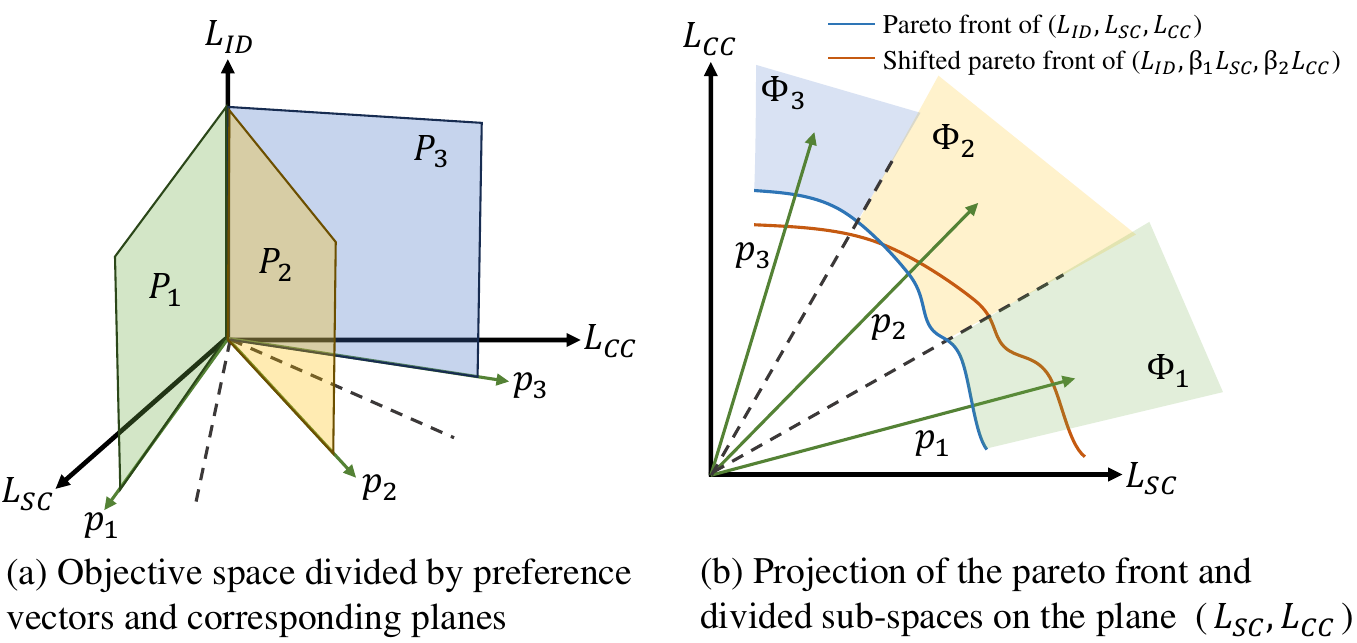}
  \caption{Illustration of optimization with preference. (a): The whole objective space is divided by a set of planes. (b): The shift of pareto front caused by directly weighting objectives in formulation~\ref{eq:opt_goal}. Our method would restrict solution to a chosen sub-space.}
  \label{fig:human_prefer}
\end{figure}
\textbf{Optimization with Human Preference.}
However, the solutions obtained by the aforementioned gradient-based optimization formulation can eventually arrive at arbitrary part on the pareto front. In real-world applications, we typically have special preference towards retrieving standard or clothes changing samples. A naive strategy to achieve this is setting fixed weights for each loss in equation~\ref{eq:opt_goal}. However, as is illustrated in Fig~\ref{fig:human_prefer}, such operation would lead to the shift of pareto front and fail to reach some original optimal solutions. To avoid this drawback, we follow the ideas~\cite{DBLP:journals/mmor/FliegeS00, DBLP:journals/tec/LiuGZ14, DBLP:conf/icml/MahapatraR20, DBLP:conf/nips/LinZ0ZK19} of dividing the objective space $\Phi$ into several sub-regions, and the optimization process would be guided to the sub-space of given preference vector.
Specifically, since the preference in CC-ReID would focus on balancing standard and CC objectives, \emph{i.e.} $\{\mathcal{L}_{sc}, \mathcal{L}_{cc}\}$ in the objective space, a set of $n$ preference vectors $\{p_1, p_2, \ldots, p_n\} \in \mathbb{R}^2$ in the plane $P_0 = \{\mathcal{L}_{id} = 0\}$ is employed to divide $\Phi$ by:
\begin{equation}
\begin{aligned}
P_k &= \{ ap_k+bu_0 \mid u_0 \perp P_0, a\in \mathbb{R}^+, b \in \mathbb{R}^+ \}, \\
\Phi_i &=\{v \in \mathbb{R}^3 \mid u_j^T v \geq u_k^T v, \forall j=1, \ldots, n\},
\end{aligned}
\end{equation}
where $u_0$ is the unit normal vector of plane $P_0$, and $u_k$ represents the unit normal vector of plane $P_k$~(see Fig~\ref{fig:human_prefer} for better understanding) with a acute angle to $v$. Therefore, $u_k$ is the normalized vector of $p_k \times u_0$. The divided objective sub-space $\Phi_i$ can be intuitively seen as the set of vectors having the smaller acute angle to the plane $P_k$ than any other preference plane. If the $k\text{-th}$ vector represents the preference of human, the MOO problem in equation~\ref{eq:moo} would be restricted by the following conditions:
\begin{equation}
\begin{aligned}
(p_j \times u_0 - p_k \times u_0)^T \hat{\mathcal{L}}(\theta) \geq 0, \forall j=1, \ldots, n.
\end{aligned}
\end{equation}
which is equivant to the constraint of
\begin{equation}
\begin{aligned}
\label{eq:restrict}
\hat{\mathcal{C}}^j(\theta)=(p_j - p_k)^T \hat{\mathcal{L}}(\theta) \leq 0, \forall j=1, \ldots, n.
\end{aligned}
\end{equation}
The optimization constrained by the restriction~\ref{eq:restrict} can be tackled by a two-stage strategy. In the first stage, the initial model parameters would be projected to the desired sub-space $\Phi_k$ in a similar way to~\ref{eq:opt_goal}:
\begin{equation}
\begin{aligned}
& \min _{\beta^1, \ldots, \beta^n} \left\|\sum_{j=1}^n \beta^j \nabla_{\theta} \hat{\mathcal{C}}^j(\theta)\right\|_2^2,
\text { s.t. } \sum_{j=1}^n &\beta^j=1, \beta^j \geq 0, \forall j \in \{j \mid \hat{\mathcal{C}}^j(\theta) \geq 0 \}.
\end{aligned}
\end{equation}
After obtaining an initial $\theta$ in the sub-space $\Phi_k$, we can re-organize the optimization formulation~\ref{eq:opt_goal} by adding dual form of sub-space constraints, and the final optimization task can be written as:
\begin{equation}
\begin{aligned}
\label{eq:final}
& \min _{\gamma^1, \ldots, \gamma^m} \left\|\sum_{i=1}^m \gamma^i \nabla_{\theta} \hat{\mathcal{L}}^i(\theta)\right\|_2^2,
\gamma^i = \alpha^i + \sum_{j=1}^n \beta^j (p_{ji} - p_{ki})\\
&\text { s.t. } \quad \sum_{i=1}^m \gamma^i=1, \quad \gamma^i \geq 0, \quad \forall j \in \{j \mid \hat{\mathcal{C}}^j(\theta) \geq 0 \}.
\end{aligned}
\end{equation}
In each iteration, we follow~\ref{eq:final} to obtain adaptive weights $\{\gamma^1,\ldots, \gamma^n\}$ to balance items in $\hat{\mathcal{L}}(\theta)$, the restricted optimization process would yield a pareto optimal solution in the sub-space related to the chosen preference $p_k$.

\section{Experiments}
\subsection{Datasets and Evaluation Protocols}
We conduct experiments on image-based PRCC~\cite{DBLP:journals/pami/YangWZ21} dataset, and video-based CCVID~\cite{DBLP:conf/cvpr/GuCMBS022} dataset. PRCC is a image-based dataset captured by 3 indoor cameras. PRCC contains 33,698 images of 221 different identities, and each person wears 2 different suits of clothes.
CCVID~\cite{DBLP:conf/cvpr/GuCMBS022} is a video-based CC-ReID dataset constructed from FVG~\cite{DBLP:conf/cvpr/ZhangT0A0WW19}, it contains 2,856 video sequences and 347,833 cropped frames of 226 different identities. The sequences span from 2017 to 2018 in terms of time, and each person wears 2-5 different kinds of outfits.

In synthesis experiments, to avoid the data leak in evaluation, CC-Diffusion model is trained only on the training set of given dataset, and these training images are augmented for clothes-varying samples with the trained CC-Diffusion. Images in test sets will not get involved in synthesis. Our decomposition and optimization methods are evaluated on test sets of both CCVID and PRCC datasets. Following previous works~\cite{DBLP:journals/pami/YangWZ21, DBLP:conf/cvpr/GuCMBS022}, the evaluation protocols are defined as follows: 
(1) \textbf{general setting} reserves all samples in test set. (2) \textbf{clothes-changing (CC)} removes those wearing same clothes. (3) \textbf{same-clothes (SC)} protocol filters out samples with different outfits. We employ Mean Average Precision (mAP) and top-1 accuarcy to measure performance for all protocols.

\subsection{Implementation Details}
For clothes-varying synthesis, both UNets in CC-Diffusion are trained from scratch for 100 epochs with batch size 32 by AdamW~\cite{loshchilov2017decoupled}, and the encoder and decoder are frozen during training. For initial 1,000 steps, the learning rate would linearly increase to 1e-4, and keeps constant until the end. The default output image size is set to $256\times128$. The corruption noise is sampled for 1,000 steps with DDPM \cite{ho2020denoising} at training, and 50 steps with DDIM \cite{song2020denoising} at inference.

For the CC-ReID learning, we employ ViT-S~\cite{DBLP:conf/iclr/DosovitskiyB0WZ21} pretrained on ImageNet~\cite{DBLP:conf/cvpr/DengDSLL009} as our default backbone model. The input image is resized to $256 \times 128$, and SGD is employed as the default optimizer with a momentum scalar set to 0.9. The model is trained for 120 epochs using the cosine learning rate scheduler. The maximum learning rate is set to 0.0016 and the learning rate period is set to 20 epochs.
To handle video sequences in CCVID dataset, we follow the original paper~\cite{DBLP:conf/cvpr/GuCMBS022} to use a ResNet-50~\cite{DBLP:conf/cvpr/HeZRS16} backbone, where spatial max pooling and temporal average pooling are employed for extracting video feature. The hyper-parameters, resolutions, and data-prepossessing are all aligned with the original paper for fair comparison. All models are trained with a NVIDIA A100 GPU.

\begin{table}[t]
\setlength{\abovecaptionskip}{2mm}
\centering
\caption{Quantitative analysis of synthetic data quality. CCSR denotes `Clothes Changing Success Rate'. CCSR metrics are not applicable for real images and pixel-sampled images by SPS, because no clothing images can be referred for metric distance.}
\begin{tabular}{p{2.4cm}|p{1.8cm}<{\centering}p{2cm}<{\centering}p{2cm}<{\centering}}
\hline
Methods             & FID$\downarrow$     & CCSR-P$\uparrow$  & CCSR-M$\uparrow$     \\ 
\hline \hline  
Real Images &\underline{12.7} &N/A &N/A \\ \hline
SPS~\cite{DBLP:journals/spl/ShuLWRT21}  &$126.8_{+114.1}$ &N/A &N/A \\ 
DG-Net~\cite{zheng2019joint}   &$26.4_{+13.7}$ &18.6  &54.8   \\ \hline
CC-Diff~(Ours)   &$\textbf{17.2}_{+4.5}$ &\textbf{84.5}  &\textbf{96.6} \\ \hline
\end{tabular}
\label{tab:fid}
\end{table}
\subsection{Ablation Studies}
\textbf{Quality of Image Synthesis.}
We first investigate the quality of synthetic clothes-changing data via quantitative analysis. 
Fréchet Inception Distance (FID) metric is applied to evaluate fidelity of synthetic images. On PRCC dataset, the FID score of synthesis by CC-Diffusion model is just higher than the real-world samples by 4.5, indicating a realistic fidelity level. Comparing to other augmentation methods in ReID, CC-Diffusion shows a better synthesis quality on clothe s-varying samples by a margin. However, FID metric can not comprehensively reflect the success or failure of the clothing-changing synthesis. Therefore, we propose a new metric CCSR~(\textbf{C}lothes \textbf{C}hanging \textbf{S}uccess \textbf{R}ate) to assist judging the fidelity of data synthesis. Specifically, CCSR employs a trained ReID model to extract features for both synthetic and original data. For each synthetic image, CCSR calculates the normalized euclidean distance between the corresponding original images~($I$ in Fig~\ref{fig:pipeline}) and clothing images~($I^{'}$ in Fig~\ref{fig:pipeline}), and one synthetic image is considered as successfully clothes changing sample if the distance with original clothing images is smaller. In Tab~\ref{tab:fid}, we report the CCSR results using the baseline model respectively trained on PRCC and MSMT17 \cite{wei2018person}, and denote it as CCSR-P and CCSR-M, respectively. We can make a clear observation that our synthesis are closer to the clothing sources for models trained on both CC-ReID and standard ReID datasets. That means models trained with existing data are not able to discriminate the high-quality synthesis. 

\begin{table}[t]
\setlength{\abovecaptionskip}{2mm}
\centering
\caption{Ablation studies on PRCC. `+Synthesis' means adding synthesis images to training data. `Only Synthesis' refers to only applying synthesis images for training excluding the original images. `+LS' and `+GBO' denotes using objective decomposition and fixed linear scalarization/ gradient-based optimization, respectively.}
\begin{tabular}{p{3.7cm}|p{0.9cm}<{\centering}p{0.9cm}<{\centering}|p{0.9cm}<{\centering}p{0.9cm}<{\centering}}
\hline
 & \multicolumn{2}{c|}{CC} & \multicolumn{2}{c}{SC}    \\ \cline{2-5}
{\multirow{-2}{*}{Methods}}        & mAP     & top-1  & mAP     & top-1 \\ 
\hline \hline
Baseline   &45.5  &45.4  &99.1  &\textbf{100} \\
Baseline-SC  &27.8 &22.3 &\textbf{99.9}  &99.8 \\
SPS \cite{DBLP:journals/spl/ShuLWRT21} &57.2  &62.8 &96.7  &99.5  \\
DG-Net \cite{zheng2019joint} &38.4  &43.8 &97.4 &99.6  \\ \hline
+Synthesis  &61.7  &69.5 &89.4  &97.0 \\ 
Only Synthesis  &59.0  &67.4 &90.2  &96.8 \\ \hline
+LS  &47.2  &49.4 &98.8  &99.8 \\
+GBO  &48.4  &50.5 &99.1  
&\textbf{100} \\ 
\hline
+Synthesis+LS  &61.9  &69.9 &93.4  &98.5 \\
+Synthesis+GBO~(Ours)  &\textbf{63.8}  &\textbf{70.1} &97.1  &99.8 \\ \hline \hline
Res50~\cite{DBLP:conf/cvpr/HeZRS16}+Ours &65.9  &71.6 &95.3 &99.8 \\
CAL~\cite{DBLP:conf/cvpr/GuCMBS022}+Ours  &\textbf{67.1}  &\textbf{73.2} &98.9  &99.9 \\ 
AIM~\cite{DBLP:conf/cvpr/YangLZW023}+Ours  &66.5  &72.9 &99.1  &\textbf{100} \\ \hline
\end{tabular}
\label{tab:prcc}
\end{table}
\textbf{Effectiveness of Clothes-Varying Synthesis.}
We perform detailed ablations for investigating training with synthetic CC data, and report the results in Tab~\ref{tab:prcc}. The baseline TransReID~\cite{DBLP:conf/iccv/He0WW0021} model achieves 45.5\% CC mAP and 99.1\% SC mAP. When SPS is employed to augment training data, we can observe a 10.6\% improvement under CC setting, and a 5.7\% degradation on SC protocol. This further validate the conflicting relationship between objectives in CC-ReID. When augmented with data generated by DG-Net, the performance on CC query sets decreases. Recalling the analysis of synthetic images quality in Tab~\ref{tab:fid}, the result here proves the significance of successful clothes-changing synthesis.
By integrating our synthetic imaged into training, 16.2\% improvement in terms of mAP can be observed under the CC protocol. As for the SC test set, a clear 9.7\% mAP drop indicates the sacrifice of standard ReID objective. Surprisingly, when the original training set is removed, the model still achieves a rather high performance. Such a result suggests that the distribution is dominated by our synthesis, and validates the diversity and quality of our synthetic data.

\begin{figure}[t]
\setlength{\abovecaptionskip}{8pt}
  \centering
    \subfloat{%
       \includegraphics[width=0.45\linewidth]{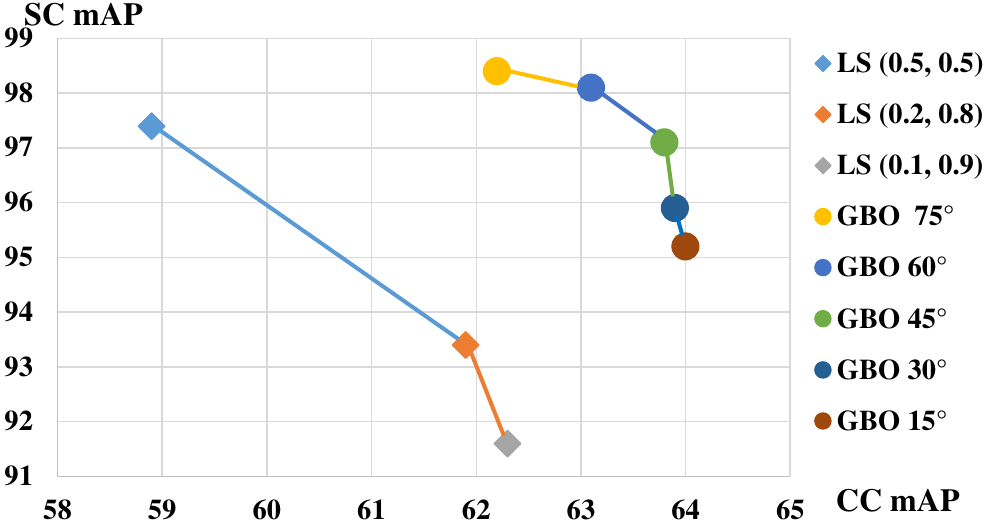}
    }
    \hspace{8mm}
    \subfloat{%
       \includegraphics[width=0.45\linewidth]{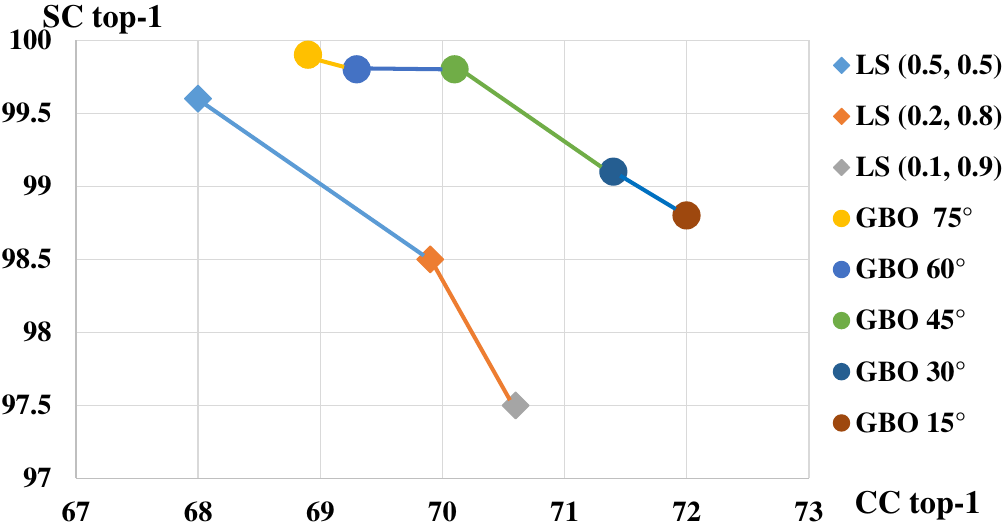}
    }
  \caption{Comparison between linear scalarization and gradient-based optimization with preference. Rhombus points denote results of linear scalarization with a fixed weights for~($\mathcal{L}_{sc}, \mathcal{L}_{cc}$). Circular points refer to GBO methods with a selected preference representing the angle in the plane $\mathcal{L}_{sc}, \mathcal{L}_{cc}$~(\emph{i.e.} GBO $45^\circ$ means the vector is ~($\frac{\sqrt{2}}{2}, \frac{\sqrt{2}}{2}$)).}
  \label{fig:preference}
\end{figure}
\textbf{Effectiveness of Objective Decomposition.} We use both original and synthetic training data, and apply the objective decomposition to the training process. We first report the results of employing manually designed linear weights~(0.5, 0.1, 0.4) to balance the decoupled objectives~($\mathcal{L}_{id}, \mathcal{L}_{sc}, \mathcal{L}_{cc}$). It can be observed that such decomposed learning process improves both CC and SC performance, proving the correctness of our proposed decomposition formulation. On such basis, the re-formulated CC-ReID problem would achieve one of the pareto optimal solutions with the gradient-based optimization method. Our ablation in Tab~\ref{tab:prcc} validates this, and the solution acquired by GBO largely outperforms the linear scalarization approach for both evaluation settings.

\textbf{Property of Model-Agnostic.}
The whole framework is model-agnostic and \textbf{complementary} to existing CC-ReID researches. Our method achieved even higher CC performance~(65.9\% mAP) with Res50~\cite{DBLP:conf/cvpr/HeZRS16}. As demonstrated in Tab~\ref{tab:prcc}, combining our method with existing works~\cite{DBLP:conf/cvpr/GuCMBS022,DBLP:conf/cvpr/YangLZW023} yields further improvements.

\textbf{Effectiveness of Preference Vector.}
The results of using LS and GBO with preference is reported in Fig~\ref{fig:preference}, where the trade-off relation between SC and CC performance can be clearly observed. With the weighting of $\mathcal{L}_{cc}$ rising from 0.5 to 0.9, the SC mAP declines from 97.4\% to 91.6\%. Fixed linear weighting does not necessarily produce optimal solution due to the non-convex pareto front. This inferiority of fixed linear weighting is clearly observable that each solution obtained by fixed linear weighting would be dominated by at least one solution of GBO. We can also make an observation that preference vector can accurately reflect human preference and steadily yield well-balanced results.

\subsection{Comparison with State-of-the-art Methods}
\begin{table}[t]
\setlength{\abovecaptionskip}{2mm}
\centering
\caption{Comparison with State-of-the-art ReID, Gait-based, and CC-ReID methods.}
\begin{tabular}{p{2.6cm}|p{0.8cm}<{\centering} p{0.8cm}<{\centering}|p{0.8cm}<{\centering} p{0.8cm}<{\centering}| p{0.75cm}<{\centering}p{0.75cm}<{\centering}|p{0.75cm}<{\centering}p{0.75cm}<{\centering}|p{0.75cm}<{\centering}p{0.75cm}<{\centering}}
\hline
 & \multicolumn{4}{c|}{PRCC} & \multicolumn{6}{c}{CCVID}    \\ \cline{2-11}
 & \multicolumn{2}{c|}{CC} & \multicolumn{2}{c|}{SC} & \multicolumn{2}{c|}{General} & \multicolumn{2}{c|}{CC} & \multicolumn{2}{c}{SC} \\ \cline{2-11}
{\multirow{-3}{*}{Methods}}  & mAP & top-1 & mAP & top-1 & mAP & top-1 & mAP & top-1 & mAP & top-1\\ 
\hline \hline
Triplet~\cite{DBLP:journals/corr/HermansBL17} &43.3  &45.6 &97.9 &99.8 &78.1  &81.5  &77.0 &81.1  &\textbf{98.8}  &\textbf{100} \\
HACNN~\cite{DBLP:conf/cvpr/LiZG18} &-  &21.8  &-  &82.5 &- &- &- &- &- &-  \\
PCB~\cite{DBLP:conf/eccv/SunZYTW18} &38.7  &41.8  &97.0  &99.8 &- &- &- &- &- &- \\
IANet~\cite{DBLP:conf/cvpr/HouMCGSC19a} &27.8 &22.3 &\textbf{99.9}  &99.8 &- &- &- &- &- &- \\
TransReID~\cite{DBLP:conf/iccv/He0WW0021} &45.5  &45.4  &99.1  &\textbf{100} &- &- &- &- &- &-  \\ \hline
GaitSet~\cite{DBLP:conf/aaai/ChaoHZF19} &- &- &- &- &72.3  &81.9  &62.1 &71.0  &-  &- \\
TCLNet~\cite{DBLP:conf/eccv/HouCMSC20} &- &- &- &- &77.9  &81.4  &75.9 &80.7  &-  &- \\
AP3D~\cite{DBLP:conf/eccv/GuCMZC20} &- &- &- &- &79.2  &80.9  &77.7 &80.1  &-  &-\\ \hline
SPS~\cite{DBLP:journals/spl/ShuLWRT21} &57.2  &62.8 &96.7  &99.5 &- &- &- &- &- &-  \\
RCSANet~\cite{DBLP:conf/iccv/00230X0Z21} &48.6  &50.2 &97.2 &100 &- &- &- &- &- &- \\
CAL~\cite{DBLP:conf/cvpr/GuCMBS022} &55.8  &55.2 &99.8 &\textbf{100} &81.3  &82.6  &79.6 &81.7  &97.9  &\textbf{100}\\
3DInvar~\cite{liu2023learning} &57.2  &56.5 &- &- &66.1  &70.8 &65.4  &70.2 &-  &- \\
3D+CAL~\cite{liu2023learning} &21.4 &40.7 &- &- &82.6  &83.9 &81.3  &84.3 &-  &- \\
AIM~\cite{DBLP:conf/cvpr/YangLZW023} &58.3  &57.9 &\textbf{99.9}  &\textbf{100} &- &- &- &- &- &-\\ 
CCFA~\cite{han2023clothing} &58.4  &61.2 &98.7  &99.6 &- &- &- &- &- &- \\
CRE+BSGA~\cite{DBLP:conf/bmvc/MuLLY22} &58.7  &61.8 &97.3  &99.6 &- &- &- &- &- &- \\
SCNet~\cite{DBLP:conf/mm/GuoLWWW23} &59.9  &61.3 &97.8  &100 &- &- &- &- &- &- \\ \hline
Ours (LS) &61.9  &69.9 &93.4  &98.5 &85.4  &88.2 &80.5  &85.7 &94.6  &99.3\\
Ours (GBO) &\textbf{63.8}  &\textbf{70.1} &97.1 &99.8 &\textbf{87.1}  &\textbf{89.7} &\textbf{83.5}  &\textbf{86.9} &97.1  &\textbf{100}\\\hline
\end{tabular}
\label{tab:sota_prcc}
\end{table}
We compare our methods with state-of-the-art standard ReID methods, and CC-ReID methods in Tab~\ref{tab:sota_prcc}. 
Our proposed method is a model-agnostic framework without any specifically designed model structure towards CC-ReID. Most CC-ReID methods put attention on introducing auxiliary modules to resist clothing variations. On the contrary, our framework uncovers a surprising finding- \textbf{a standard ReID model structure with high-fidelity synthetic CC data and MOO formulation can largely outperforms model-centric CC-ReID methods}.
Moreover, we can observe a similar degradation under the SC protocol in some feature augmentation methods~\cite{han2023clothing,DBLP:journals/spl/ShuLWRT21}, which indicates the universality of this concealed conlficts in CC-ReID objectives. On PRCC, our framework finally takes an advantage of over 7\% on CC top-1 accuracy, with similar SC performance to SoTA methods.
On the video-based CCVID dataset, we compare our framework with video ReID~\cite{DBLP:conf/eccv/HouCMSC20,DBLP:conf/eccv/GuCMZC20} methods, gait-based methods~\cite{DBLP:conf/aaai/ChaoHZF19}, and video CC-ReID methods. Our method suppresses SOTA method by 4.5\% in terms of general mAP. Such improvements not only illustrate the high-fidelity and \textbf{temporal consistency} of synthetic video frames, but also reflect the effectiveness of our MOO formulation on video scenarios. Results on more CC-ReID datasets are presented in Appendix.


\section{Conclusion}
In this paper, we first uncover the conflicting relation between standard and CC learning objectives inside CC-ReID, and propose to generate high-fidelity clothes-changing synthesis for boosting CC performance. We mitigate such inner conflicts through objective decomposition and re-formulating CC-ReID as a multi-objective optimization problem, and a gradient-based method with human preference is applied to yield desired pareto optimal solutions. We hope this work will inspire more exploration towards the complex learning process of CC-ReID.


%
%
\bibliographystyle{splncs04}
\bibliography{main}
\end{document}